\documentclass[twoside,twocolumn,10pt]{article}

\usepackage{wscg}           
\RequirePackage{ifpdf}
\ifpdf
 \RequirePackage[pdftex]{graphicx}
 \RequirePackage[pdftex]{color}
\else
 \RequirePackage[dvips,draft]{graphicx}
 \RequirePackage[dvips]{color}
\fi

\usepackage{amsmath}
\usepackage{calc}
\title{Leveraging Large Language Models to Effectively Generate Visual Data for Canine Musculoskeletal Diagnoses}

\author{
\parbox{\textwidth}{\centering
Martin Thi{\ss}en\textsuperscript{1,3} \quad
Thi Ngoc Diep Tran\textsuperscript{1,3} \quad
Barbara Esteve Ratsch\textsuperscript{2} \quad
Ben Joel Sch{\"o}nbein\textsuperscript{1,3} \\[3mm]
Ute Trapp\textsuperscript{1,3} \quad
Beate Egner \textsuperscript{2} \quad
Romana Piat\textsuperscript{1,3} \quad
Elke Hergenr{\"o}ther\textsuperscript{1,3} \\[4mm]
\textsuperscript{1} Darmstadt University of Applied Sciences, Schöfferstraße 3, 64295 Darmstadt, Germany. \\
\textsuperscript{2} Veterinary Academy of Higher Learning, Im Riemen 27, 64832 Babenhausen, Germany. \\
\textsuperscript{3} European University of Technology, European Union.
}
}

\usepackage{url}
\makeatletter
\def\Uslash{\mathbin{\mathchar`\/}\@ifnextchar{/}{\kern-.15em}{}}
\g@addto@macro\UrlSpecials{\do \/ {\Uslash}}
\def\Ucolon{\mathbin{\mathchar`:}\@ifnextchar{/}{\kern-.1em}{}}
\g@addto@macro\UrlSpecials{\do : {\Ucolon}}
\makeatother

\begin{document}

\twocolumn[{\csname @twocolumnfalse\endcsname

\maketitle  

\begin{abstract}
\noindent
It is well-established that more data generally improves AI model performance. However, data collection can be challenging for certain tasks due to the rarity of occurrences or high costs. These challenges are evident in our use case, where we apply AI models to a novel approach for visually documenting the musculoskeletal condition of dogs. Here, abnormalities are marked as colored strokes on a body map of the dog. Since these strokes correspond to distinct muscles or joints, they can be mapped to the textual domain in which large language models (LLMs) operate. LLMs have recently demonstrated impressive capabilities across a wide range of tasks, including medical applications, offering promising potential for generating synthetic training data. In this work, we investigate whether LLMs can effectively generate synthetic visual training data for canine musculoskeletal diagnoses. For this, we developed a mapping that segments visual documentations into over 200 labeled regions representing muscles or joints. Using techniques like guided decoding, chain-of-thought reasoning, and few-shot prompting, we generated 1,000 synthetic visual documentations for patellar luxation (kneecap dislocation) diagnosis, the diagnosis for which we have the most real-world data. Our analysis shows that the generated documentations are sensitive to location and severity of the diagnosis while remaining independent of the dog's sex. However, they were also independent of the patient’s weight and age, highlighting some limitations. We further generated 1,000 visual documentations for various other diagnoses to create a binary classification dataset. A model trained solely on this synthetic data achieved an F1 score of 88\% on 70 real-world documentations. These results demonstrate the potential of LLM-generated synthetic data, which is particularly valuable for addressing data scarcity in rare diseases or rare instances in general. While our methodology is tailored to the medical domain, the insights and techniques can be adapted to other fields.
\end{abstract}

\subsection*{Keywords}
Generative AI, Large Language Models (LLMs), Synthetic Data for Veterinary Medicine

\vspace*{1.0\baselineskip}
}]


\section{Introduction}
\label{sec:intro}

The success of artificial intelligence (AI) models in solving complex tasks often depends on the availability of large, high-quality datasets. However, in domains with limited data, the performance of these models is typically constrained. Large language models (LLMs) have recently introduced novel capabilities, such as multitask learning \cite{rad19} and in-context learning \cite{bro20}. State-of-the-art LLMs \cite{ach23} \cite{gra24} \cite{ani23} \cite{guo25}, trained on large and diverse datasets, possess extensive general knowledge across numerous domains \cite{wan24} \cite{rei24}. This raises a compelling question: can LLMs be effectively used to generate synthetic training data for specialized tasks with limited initial data?
Previous studies \cite{sin24} \cite{gao23} have shown that LLMs can successfully generate synthetic data for tasks they are already familiar with from their training data. Technically, LLMs can even be trained entirely on synthetic data generated by another LLM and still achieve strong performance \cite{abd24}. However, less is understood about their ability to create high-quality synthetic data for tasks outside their training distribution, particularly for highly specialized or low-resource domains.

\begin{figure}[t]
    \centerline{\includegraphics[width=\linewidth]{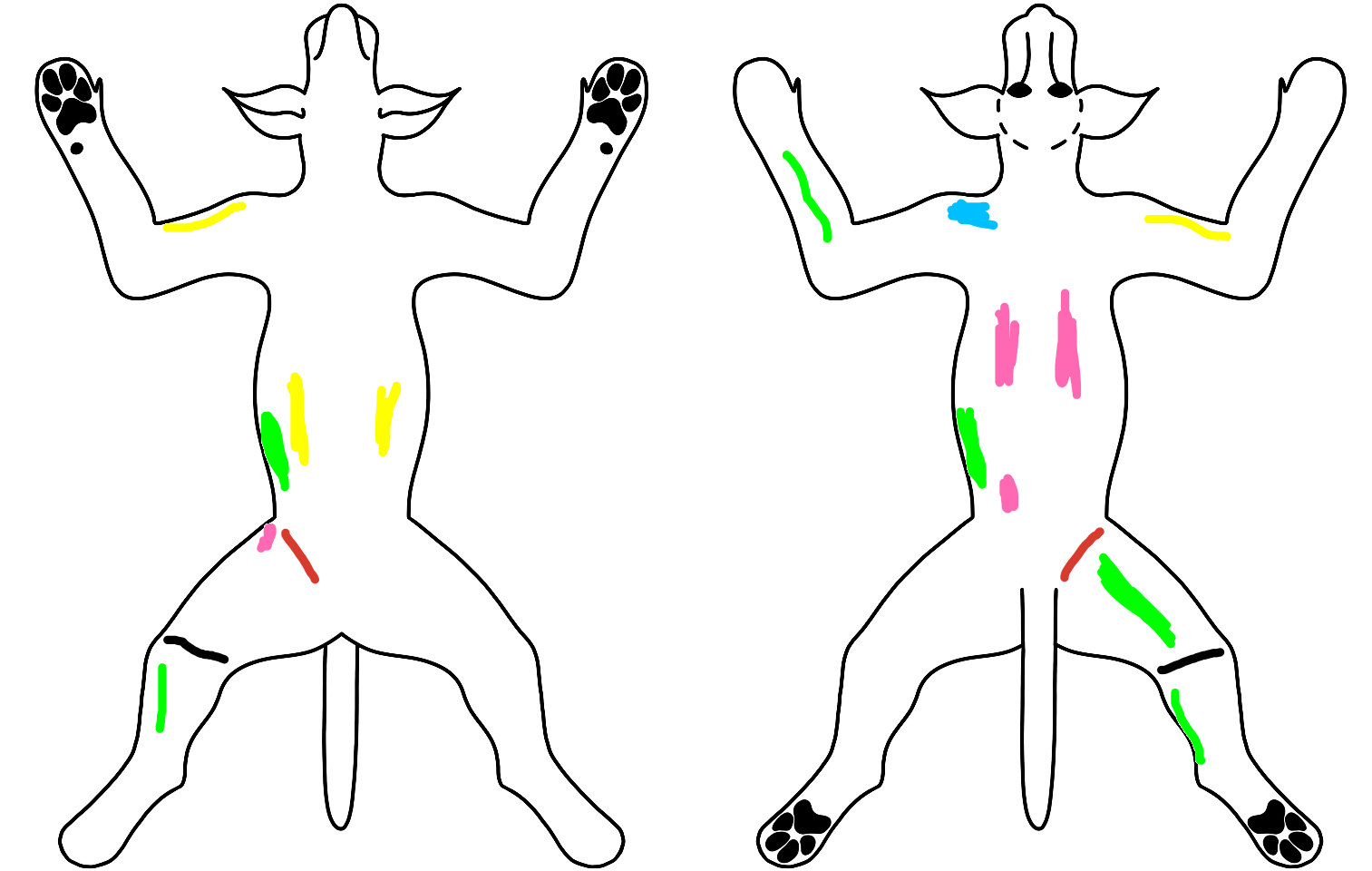}}
    \caption{An example visual documentation of a dog's musculoskeletal condition, diagnosed with patellar luxation (kneecap dislocation) using the method proposed by Esteve et al. \cite{est24}.}
    \label{fig_bodymap}
\end{figure}

In this work, we explore this potential in the context of a novel method \cite{est24} \cite{thi24} for visually documenting the musculoskeletal condition in dogs. Figure \ref{fig_bodymap} presents an example of such documentation. The colored strokes indicate abnormalities (e.g., pain or tension) in a dog's muscles and joints, detected by a veterinarian through palpation (pressing on specific areas to identify abnormalities) and documented visually using a mouse, finger, or digital pen. Specifically, strokes can be drawn within the dog illustrations in Figure \ref{fig_bodymap} using seven different colors, each representing a distinct abnormality (e.g., pain or tension). Since these strokes represent discrete anatomical entities such as muscles or joints, they can be mapped to corresponding textual descriptions. As LLMs operate in the textual domain, they can interpret visual documentation like Figure \ref{fig_bodymap} when mapped to corresponding textual descriptions. However, this requires a sufficiently detailed mapping. Additionally, with such a mapping, textual descriptions generated by an LLM could be linked to specific regions within the dog illustrations in Figure \ref{fig_bodymap}, with strokes rendered within those regions. As a result, by generating a list of textual descriptions related to a specific diagnosis using an LLM, we can create synthetic visual documentations for that diagnosis.

As this novel method \cite{est24} \cite{thi24} is newly introduced, available data is extremely limited, with only 70 real-world examples. Given that LLMs have reached or exceeded human-level performance across various medical benchmarks \cite{jin21} \cite{saa24}, they present a promising solution for generating effective synthetic data in this context. Thus, this work investigates whether LLMs can effectively generate synthetic data for canine musculoskeletal diagnosis, despite their limited prior exposure to this specific domain.

To this end, we developed a mapping between visual documentations and over 200 labeled anatomical regions. This mapping enables the LLM to interpret visual documentations and, when combined with a rendering approach that mimics human hand-drawing, which we also developed, it allows the LLM to generate visual documentations. Using our approach proposed in this work, which leverages the extensive medical knowledge and strong reasoning capabilities of state-of-the art LLMs, we generated 1,000 synthetic visual documentations of patellar luxation (kneecap dislocation) diagnosis, the diagnosis for which we have the most real-world data. Our analysis showed that the generated documentations demonstrate faithfulness and controllability, as they are sensitive to the specified location and severity of the patellar luxation while remaining independent of the dog's sex.

Building on this, we generated an additional 1,000 visual documentations across various other diagnoses, creating a binary classification dataset to determine whether a documentation represents a patellar luxation or not. We chose patellar luxation as our target diagnosis because it is the one for which we have the most real-world data. A classification model trained solely on this visual synthetic dataset achieved an F1 score of 88\% when evaluated on 70 real-world documentations, demonstrating the effectiveness, faithfulness, and diversity of the generated data.

While using an LLM to classify visual documentations directly is a promising alternative to generating synthetic training data, practical constraints must be considered. In some veterinary practices in Germany, the computer running the practice management software is intentionally kept offline to avoid security risks, requiring AI models to run locally. By leveraging synthetic data generation, we distilled the LLM's medical knowledge into a classification model that has 8,974 times fewer parameters than the LLM we used but still achieves strong performance (88\% F1 score) while remaining suitable for deployment on local hardware.

Our findings underscore the potential of using LLMs to generate synthetic training data, which is particularly valuable for rare diseases or other instances where real-world data is limited. While our approach is tailored to the medical domain, the techniques and insights can be extended to other fields.

\section{Related Work}
\textbf{Text-Generation Strategies:} 
To generate synthetic data using an LLM, the model must be provided with sufficient context. This context is conveyed through a carefully designed prompt that includes all relevant information needed to produce accurate and contextually relevant outputs. Designing such prompts often requires an explorative process known as prompt engineering, in which the goal is to iteratively refine the prompt to elicit the desired response from the model. Beyond prompt engineering, several advanced strategies have been developed to enhance the quality of LLM outputs. Holtzmann et al. \cite{hol20} introduced a decoding approach that generates text more closely resembling human-like writing, improving naturalness and coherence. Brown et al. \cite{bro20} pioneered the concept of in-context learning, where providing a few task-specific examples in the prompt enables the LLM to generalize and respond effectively to similar queries. While in-context learning achieves strong results, LLMs often struggle with tasks requiring complex reasoning. To address this, Wei et al. \cite{wei22} proposed the chain-of-thought approach, which breaks down problems into intermediate reasoning steps. This technique can be further optimized using self-consistency \cite{wan23}, where multiple candidate outputs are generated, and the most consistent answer is selected. As demonstrated by Nori et al. \cite{nor23}, these strategies have also been shown to be effective in the medical domain. While these techniques typically need to be explicitly programmed or incorporated into the input prompt, Guo et al. \cite{guo25} demonstrated that fine-tuning pre-trained LLMs in a reinforcement learning environment enables reasoning capabilities to emerge autonomously, achieving state-of-the-art performance. However, these advanced strategies come at the cost of increased computational demand, leading to longer generation times within fixed compute budgets and a larger carbon footprint.

\textbf{LLM-Generated Synthetic Data:}
As highlighted by Long et al. \cite{lon24}, two challenges commonly arise when generating synthetic data using LLMs: faithfulness and diversity. Faithfulness refers to the logical and grammatical coherence of the generated data, along with ensuring it remains free of factual errors or inaccurate labels \cite{lon24}. Long et al. \cite{lon24} note that maintaining faithfulness becomes increasingly difficult when generating long, complex, or domain-specific data. All of these characteristics are present in our scenario, making it particularly challenging. Therefore, ensuring faithfulness requires a carefully designed data generation process paired with thorough evaluation methods. Data diversity refers to the variation in generated data, such as differences in text length or writing style. Long et al. \cite{lon24} observe that uncontrolled data generation often results in repetitive and monotonous outputs, which could reduce its effectiveness as training data in our context. Consequently, it is crucial to apply strategies that address the issue of monotonous data generation while preserving faithfulness. A third key factor relevant to our use case is controllability in synthetic data generation. Specifically, since our goal is to generate synthetic data for specific diagnoses given a patient's metadata (e.g. age, sex, and weight), a high degree of controllability is essential. In alignment with Liu et al.'s \cite{liu24} recommendation for future research -- "Future research should focus on developing new advanced techniques [...] that can control and manipulate specific attributes of the generated data, enabling the creation of diverse and customizable synthetic datasets" -- our work explores this challenge within the context of canine musculoskeletal diagnosis.

\section{Visual Documentations}
In this section, we would like to provide further detail to the visual documentations shown in Figure \ref{fig_bodymap}. Diagnosing a dog's physical health after injuries, surgeries, or other conditions affecting movement typically involves a physical examination, including palpation, which is the process of feeling or pressing on specific areas of the dog's body to detect abnormalities such as pain, swelling, or muscle tension. The found abnormalities are often recorded as written notes. However, the novel approach by Esteve et al. \cite{est24} \cite{thi24} allows to document such abnormalities visually using a mouse, finger, or digital pen. This allows for a faster examination, which is especially helpful when dogs in pain are less likely to cooperate. Each visual documentation follows a consistent format: the left-side dog represents a face-up (supine) view, while the right-side dog represents a face-down (prone) view. These views are symmetrically mirrored, meaning that strokes on the right side of the left dog correspond to the left side of the actual physical dog. In the visual documentation proposed by Esteve et al. \cite{est24}, abnormalities are represented as strokes in seven distinct colors, each corresponding to different conditions such as pain or tension. Since these strokes are associated to discrete anatomical entities such as muscles or joints, they can be mapped to corresponding textual descriptions. When strokes within a visual documentation like Figure \ref{fig_bodymap} are mapped to corresponding textual descriptions, they can be interpreted by LLMs, as LLMs operate in the textual domain. However, this requires a sufficiently detailed mapping. Additionally, such a mapping allows to link textual descriptions generated by an LLM to specific regions within the dog illustrations in Figure \ref{fig_bodymap}. Within those regions, strokes could be rendered to create a visual documentation based on textual descriptions generated by an LLM. For this process to work, the generated textual descriptions of abnormalities must be specific enough to be linked to discrete anatomical entities, such as muscles or joints, and include information about the specific condition, such as pain or tension. This ensures that the right color is chosen and that the abnormality is rendered in the correct location. As a result, by generating a list of such textual descriptions for a specific diagnosis using an LLM, we would be able to create synthetic visual documentations for that diagnosis. In the following section, we present our method for generating textual descriptions of abnormalities using an LLM and transforming them into visual documentations through a mapping and rendering process.

\section{Method}
\subsection{Discretizing Visual Documentations}
\label{sec:mapping}
\begin{figure}[t]
    \centerline{\includegraphics[width=\linewidth]{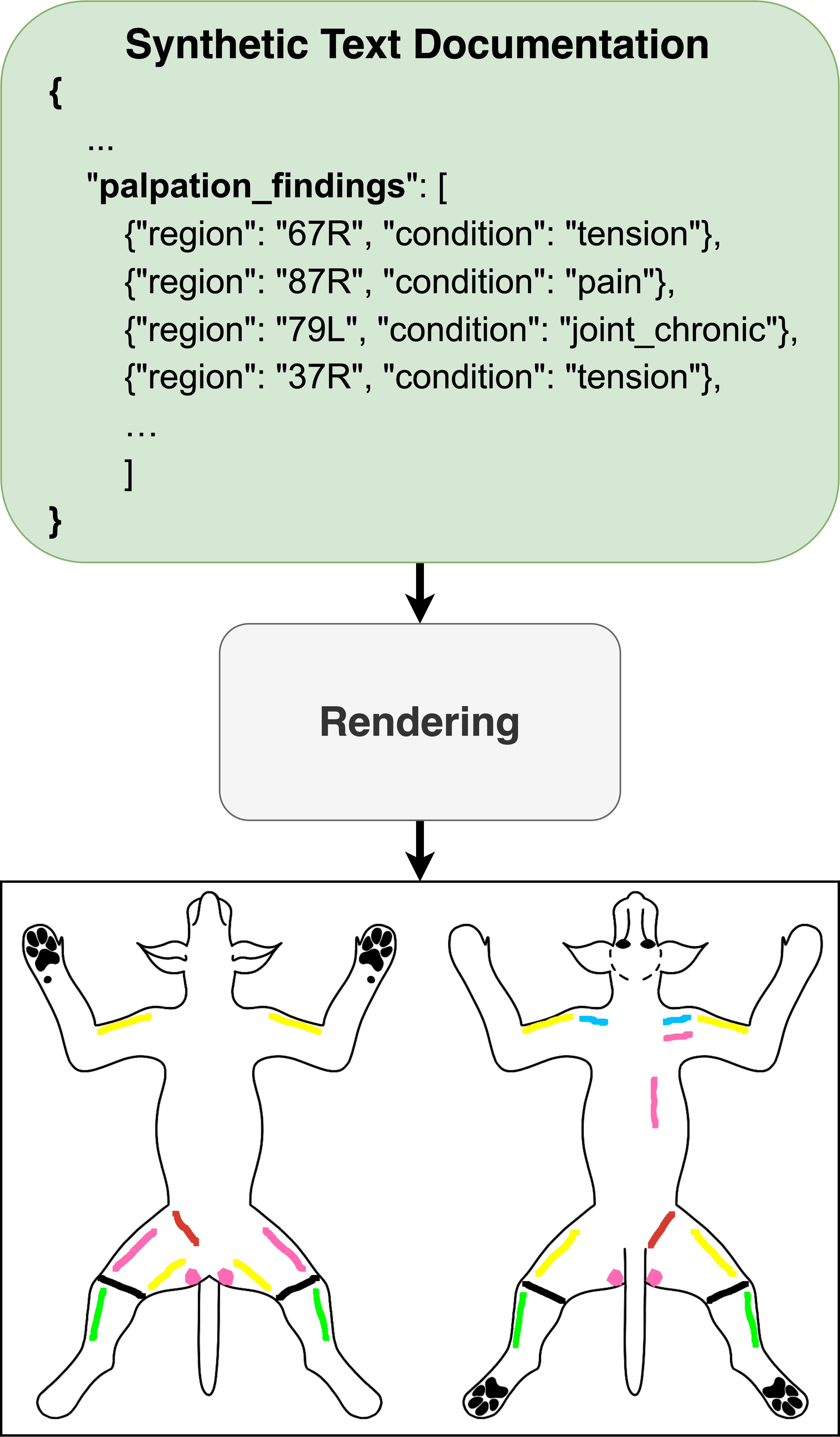}}
    \caption{Illustration of how the textual \texttt{<region index, condition index>} pairs are transformed into visual representations, allowing LLMs to generate structured visual documentations.}
    \label{fig:rendering}
\end{figure}

A novel method introduced by Esteve et al. \cite{est24} \cite{thi24} enables veterinarians to visually document a dog's musculoskeletal condition, as illustrated in Figure \ref{fig_bodymap}. These visual documentations resemble sketches composed of numerous strokes, making it challenging for large language models (LLMs) to directly generate or interpret them. To overcome this limitation, we developed a mapping that separates the visual documentation into 214 distinct regions. Each region is assigned a unique index and label, linking individual strokes to the corresponding textual descriptions of muscles, muscle groups, or joints. These regions are either rectangular or circular in shape, with most being rectangular. A combination of both shapes is used to maximize coverage within the visual documentation while ensuring that regions proportionally represent muscles or joints. Circular regions are used to represent important small soft tissue structures or muscle origins. Examples of corresponding lines in rectangular regions and circles in circular regions can be seen in Figure \ref{fig:rendering}. Similarly to the regions, the seven possible abnormality conditions, such as pain or tension, are each assigned an index and a corresponding textual description. Using this mapping of regions and conditions, textual descriptions can be assigned a corresponding color and region in the visual documentation. However, to add strokes to a visual representation, a rendering process is required. Using the region and condition indices, this process renders strokes in the designated region with the corresponding color within the visual documentation. This allows textual descriptions generated by an LLM to be transformed into a visual documentation. Our developed rendering process is described in Section \ref{sec:rendering}.

\subsection{Rendering Visual Documentations}
\label{sec:rendering}
Using the mapping introduced in the previous section, textual descriptions can be assigned a corresponding color and region index. For rendering, each abnormality requires a region index (one of 214) and a condition index (one of 7). Given a list of \texttt{<region index, condition index>} pairs, we first determine whether the corresponding region is circular or rectangular in shape. Based on this, we apply one of the following two rendering approaches. The color is chosen according to the condition index. Examples of both lines and circles are shown in Figure \ref{fig:rendering}.

\textbf{Lines.} For rectangular regions, we first determine the start point $s$ and end point $e$ based on the primary orientation of the region's contour. These points are then randomly shifted in both the x- and y-direction within a range of $[-5, 5]$ pixels, resulting in the adjusted start point $\hat{s}$ and end point $\hat{e}$. Using this pair of adjusted points, a hand-drawn mimicked line is rendered using the algorithm proposed by Meraj et al. \cite{mer08}.

\textbf{Circles.} For circular regions, we render a hand-drawn mimicked circle using a custom implementation. The process begins by identifying the center point $c$ and radius $r$ of the circular region. Next, eight control points are obtained evenly distributed along the circumference of the circle. These control points are then each shifted by a random value within the range of $[-0.15r, \; 0.15r]$ pixels in both the x- and y-direction. The shifted points form a polygon, which is rendered as a filled shape to create a hand-drawn mimicked circle.

\subsection{Metadata Knowledge Base}
\label{sec:metadata}
In our initial attempts to generate synthetic data using less specific prompts, we encountered limitations where the LLM (Llama 3 70B Instruct model \cite{gra24}) consistently produced a narrow range of outputs for patient metadata, such as only five different dog breeds and ages limited to 3 or 4 years. However, since our primary goal is to generate visual documentation as training data, we decided to simplify the task for the LLM. Instead of asking it to generate both the synthetic documentation and metadata for each dog, we decided to provide the metadata as input, allowing the LLM to focus solely on creating the documentation. This approach not only reduces the task complexity but also ensures the creation of a balanced synthetic dataset with respect to patient metadata. To achieve this, we extracted a canine metadata knowledge base from Wikipedia, which includes 149 dog breeds along with gender-specific information on life expectancy, minimum weight, and maximum weight. This knowledge base enables us to uniformly sample synthetic patient metadata, including breed, age, sex, and weight. For each synthetic documentation generated, we incorporate newly sampled metadata into the LLM prompt. This way, we can systematically generate balanced and diverse synthetic data.

\subsection{Generating Textual Documentations}
\label{sec:generation}
Our main objective of generating complex and domain-specific synthetic data presents a significant challenge, as it is difficult to reliably elicit such outputs from LLMs. Through a series of controlled experiments, we developed a well-structured prompt to reliably guide the LLM in generating the intended synthetic data. In total, the prompt contains 1,500 words or 13,918 characters. The prompt is structured into the following six components:

\begin{enumerate}
    \item \textbf{Task Definition and Background:} We begin by stating the task and providing brief background information on how the data is typically collected (through palpation by a veterinarian).
    \item \textbf{Condition Descriptions:} We list all seven possible conditions, including their indices and descriptions.
    \item \textbf{Region Descriptions:} We provide a detailed list of 214 regions, including their indices and descriptions.
    \item \textbf{Example Documentations:} We include four real-world patellar luxation documentations (grades 1–4), mapped to text using the mapping described in Section~\ref{sec:mapping}. The number of palpated abnormalities in these examples ranges from 9 to 51.
    \item \textbf{Instruction Set:} We specify key constraints to ensure high-quality, relevant synthetic data, such as:
    \begin{itemize}
        \item ``Each region can only have one abnormality (condition).''
        \item ``Abnormalities are either present or absent, not mild or severe.''
    \end{itemize}
    \item \textbf{Dynamically Sampled Patient Metadata:} At the end of the prompt, we append sampled textual metadata about the patient (see Section~\ref{sec:metadata}). This is the only dynamic component of the prompt.
\end{enumerate}

\vspace{-2mm}

In the following, we highlight key properties of our data generation approach using this prompt.

\textbf{Few-Shot In-Context Learning.} 
Initially, we observed that the generated documentations sometimes did not align with the expected style and wording. For instance, instead of specifying individual muscles, the model occasionally generated ambiguous regions, such as the "right hindlimb", which are too vague to be mapped to one of our 214 predefined regions. Moreover, a single abnormality was sometimes assigned multiple conditions (e.g., pain and tension simultaneously), complicating the process of separating them into distinct abnormalities during post-processing. To address these issues, we incorporated four real-world visual documentations into the prompt. These documentations were converted to text using the mapping introduced in Section~\ref{sec:mapping}, providing the model with concrete examples of the desired output format.

\textbf{Control of Specific Attributes.} As outlined in Section~\ref{sec:metadata}, we constructed a canine metadata knowledge base. Before generating a synthetic documentation, we randomly sample a patient's metadata from this knowledge base and incorporate it into the prompt. This metadata includes details such as the dog's breed, age, sex, and weight, allowing us to create a balanced and diverse synthetic dataset.

\textbf{JSON-formatted Responses.}
As it is challenging to elicit outputs from LLMs in a standardized format that can be programmatically processed, we applied guided decoding \cite{wil23, don24} to ensure that the LLM generates outputs as valid JSON objects, which can be processed to generate visual documentations as shown in Section~\ref{sec:rendering}. Guided decoding \cite{wil23, don24} enables precise control over an LLM’s output by dynamically restricting the possible next tokens at each generation step based on a predefined structure, such as a JSON schema or a regular expression.

\textbf{Chain-of-Thought Reasoning.}
Chain-of-thought reasoning has been shown to improve the performance of LLMs, either explicitly by instructing them to reason step-by-step within the prompt \cite{wei22} or implicitly through newer reasoning models such as DeepSeek R1 \cite{guo25}. Guided decoding, however, can constrain the reasoning abilities of LLMs \cite{tam24}. To mitigate this, we divide the generation process into three distinct steps, leveraging the benefits of both approaches. This also simplifies the task compared to generating a synthetic documentation as a JSON object in a single pass. The required JSON object includes the patient's metadata and a list of palpated abnormalities based on the given diagnosis. Generating this in one step would require the LLM to simultaneously reason about plausible musculoskeletal anomalies and recall the corresponding region and condition indices from 214 distinct regions and seven condition types. The three distinct steps are described below.

\subsubsection{Freeform Generation}
During the first step, using the prompt shown above, the LLM generates freeform text, initially reasoning about how a synthetic documentation for a given patient (metadata and diagnosis) should be structured, and then producing the synthetic documentation as a result of this reasoning.

\subsubsection{JSON Object Conversion}
Once the reasoning pass is complete, we extract the generated freeform synthetic documentation using a reasoning delimiter (e.g., \texttt{</think>} in the case of DeepSeek R1). For explicitly instructed chain-of-thought reasoning, we recommend instructing the LLM to enclose its reasoning within \texttt{<think>} and \texttt{</think>} delimiters. This ensures compatibility with our implementation, even for non-reasoning models. We then apply guided decoding \cite{wil23} \cite{don24}, as discussed earlier, to convert the extracted freeform documentation into a standardized, typed JSON object. To achieve this, we needed two components: a defined schema for the desired JSON object format and a prompt instructing the model to convert the given input text into a JSON object that adheres to this schema. Our defined schema includes all metadata (breed, age, sex, and weight) as well as a list of strings called \textit{palpation\_findings}, which will later be rendered as strokes in the visual documentations (see Figure~\ref{fig:rendering}). The prompt provides brief background information about the given input text and instructs the model to convert it into a valid JSON object according to the given schema. The prompt consists of 59 words, two of which are placeholders for the extracted freeform documentation and the expected JSON output schema, which can be dynamically replaced. For each extracted freeform documentation, this prompt is dynamically filled, and a JSON object is generated using guided decoding \cite{don24}, which ensures that the LLM generates only tokens that form a valid JSON object according to our defined schema.

\subsubsection{Discretizing Abnormalities}
Now that the synthetic documentation has been converted into a JSON object, one final step remains before it can be rendered as a visual documentation (see Section~\ref{sec:rendering}). The abnormalities stored in the \textit{palpation\_findings} list within the JSON object are still in freeform text. However, to allow rendering, these must be mapped to discrete region and condition indices, as illustrated in Figure~\ref{fig:rendering}. To achieve this, we incorporated a list of all 214 distinct regions, each represented as a \texttt{<region index, region label>} pair, as well as all seven \texttt{<condition index, condition label>} pairs into a new prompt. At the end of the prompt, we instruct the LLM to determine the region and condition index that best matches a given abnormality description. The prompt also contains a placeholder for the abnormality description at the end, which can be dynamically filled. Furthermore, we defined a schema with two fields: one for the region and one for the condition. The region field is constrained by a regular expression, allowing only one of the 214 valid region indices, while the condition field is defined as an enumerated type with seven possible values corresponding to the condition indices. Using this schema and prompt, we apply guided decoding \cite{wil23} to ensure that the LLM generates only valid region and condition indices. This approach enables us to systematically convert all freeform abnormalities into their corresponding region and condition indices.

\subsection{Generation Setup}
To generate synthetic text documentations, we used the DeepSeek R1 Distill Llama 70B model \cite{guo25} with AWQ quantization \cite{lin24} to accommodate our limited computational resources (1x NVIDIA RTX 6000 Ada GPU). Since dynamic inputs were added at the end for all three used prompts, we used prefix caching \cite{jha23} to improve generation efficiency. Following Guo et al. \cite{guo25}, we set the generation temperature to 0.6 and applied a top-p threshold \cite{hol20} of 0.95. The generation temperature controls the randomness of the model's outputs, with lower values (closer to 0) making outputs more deterministic and higher values (closer to 1) increasing diversity. The top-p threshold \cite{hol20} controls output diversity by restricting token selection to the most probable ones until their cumulative probability reaches p (0.95 in our case), reducing randomness while maintaining coherence. Using this configuration and the three-step setup described in the previous section, we generated 1,000 synthetic documentations for the diagnosis of patellar luxation (kneecap dislocation), a condition for which we have the most real-world data.

\section{Results}

\begin{figure*}[t]
\centerline{\includegraphics[width=\linewidth]{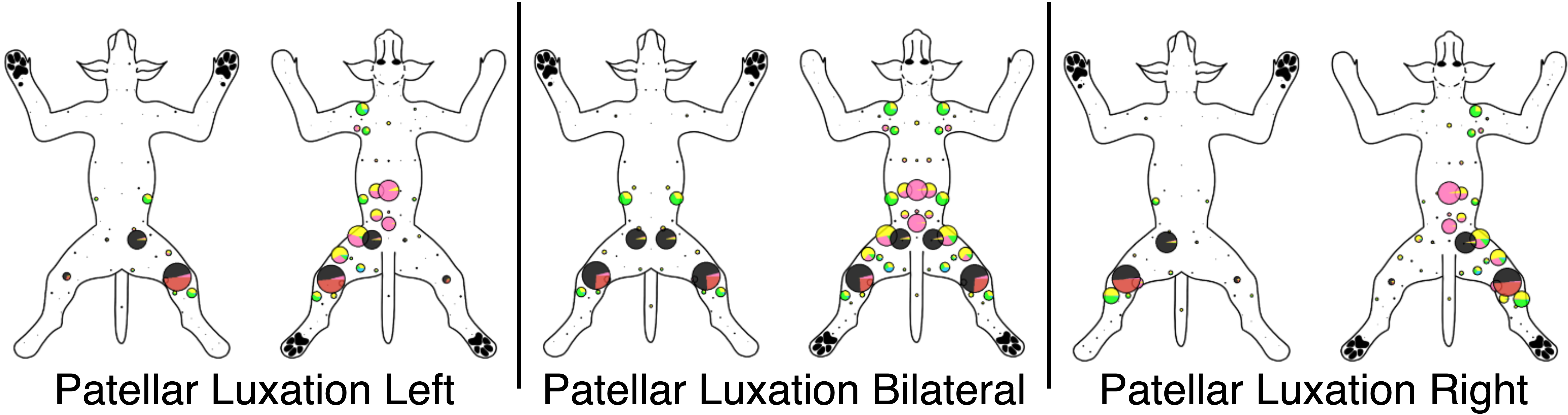}}
    \caption{The size of an individual bubble indicates the relative frequency of strokes in the corresponding region. The colors within the bubble indicate the relative frequency of a condition (e.g. pain) in the respective region. As can be seen, the LLM is able to generate synthetic documentations of a patellar luxation that represent the given positional information (left, bilateral, and right).}
    \label{fig:bodymap_location_awareness}
\end{figure*}

After generating 1,000 visual documentations for the patellar luxation diagnosis, we aimed to evaluate their faithfulness, diversity, and controllability. While manual assessment by domain experts would be the ideal approach, it is both labor-intensive and cost-prohibitive, particularly when scaling to larger volumes of generated documentations or extending to additional diagnoses. To overcome this challenge, we developed alternative measures to analyze the generated data. These measures allow to evaluate across different diagnoses and larger datasets with minimal additional effort. Crucially, they were designed in collaboration with domain experts (veterinarians) to ensure their relevance and helpfulness.

\subsection{Awareness of Severity and Position}
\label{sec_awareness}
\begin{figure}[t]
\centerline{\includegraphics[width=\linewidth]{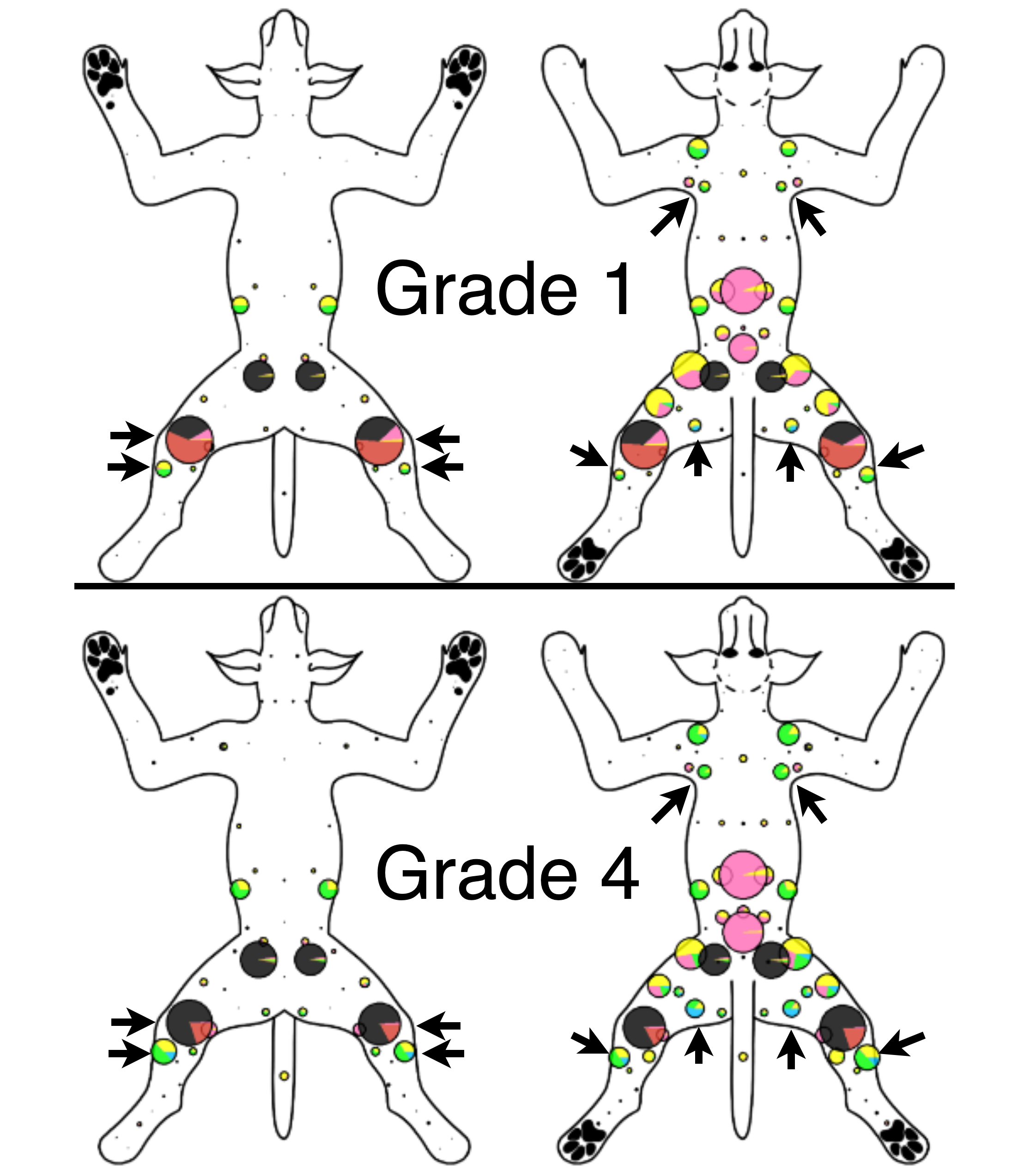}}
    \caption{Comparison of stroke distribution in patellar luxation grades 1 and 4. Arrows highlight key differences, including increased abnormalities and a regional shift in the knee joints, where acute inflammation (red) in grade 1 progresses to chronic damage (black) in grade 4.}
    \label{fig:bodymap_grade_awareness_v2}
\end{figure}

We first developed a measure to easily observe the overall distribution of visual strokes for the generated synthetic documentations. Specifically, we counted the frequency of strokes across all regions and the frequency of conditions within specific regions. Using this data, we created a bubble chart visualization to provide an intuitive overview of the structure of the generated data. Examples of these visualizations are shown in Figure \ref{fig:bodymap_location_awareness}, where we compared the distribution of strokes based on the location of the patellar luxation (left, bilateral, right). The location information was passed to the LLM as part of the input prompt. To interpret the visualization and the visual documentations accurately, it is important to note that in each visual documentation, the dog on the left represents the face-up view (supine), while the dog on the right represents the face-down view (prone). These views are symmetrically mirrored, meaning strokes on the right side of the left dog correspond to the left side of the actual physical dog. Using this approach, we found that the LLM successfully generated synthetic visual documentations that accurately reflect the location of the patellar luxation, as shown in Figure \ref{fig:bodymap_location_awareness}. We extended this strategy to evaluate the severity of patellar luxation (grades 1-4). A full comparison across all grades is provided in the appendix (Figure~\ref{fig:bodymap_grade_awareness}), while Figure~\ref{fig:bodymap_grade_awareness_v2} highlights key differences between grades 1 and 4. These comparisons illustrate that higher grades of patellar luxation correspond to an increased number of strokes (abnormalities) generated by the LLM. The average number of abnormalities for each grade of patellar luxation is as follows: 11.52 for grade 1, 13.4 for grade 2, 15.22 for grade 3, and 16.95 for grade 4. This trend is particularly evident when comparing grade 1 to grade 4 patellar luxations (see Figure~\ref{fig:bodymap_grade_awareness_v2}). Notably, healthy dogs typically have few or no abnormalities, indicating that the LLM effectively incorporated grade-specific information into the synthetic visual documentations. Additionally, regional shifts in abnormalities can be observed, particularly in the knee joints. For instance, in grade 1, most abnormalities are colored red, indicating acute inflammation, whereas in grade 4, they are predominantly black, signifying chronic joint damage.

\subsection{Impact of Controllable Attributes}
After analyzing diagnosis-specific attributes in the previous section, we extended our investigation to assess how the metadata attributes -- age, sex, and weight -- provided in the input prompt influence the generated data. Our approach for analyzing each attribute is detailed below.

\textbf{Age.} To analyze the impact of age, we first examined the age distribution across all 1,000 generated documentations. We then divided the documentations into three approximately equal-sized bins: the first bin included ages 1–5 years, the second spanned ages 6–9 years, and the third covered ages 10–20 years. Using the same evaluation method described in Section~\ref{sec_awareness}, we evaluated the generated data for each bin. The bubble charts for all three age bins, as well as the average number of abnormalities, are nearly identical, suggesting that age does not significantly influence the generation of synthetic documentations. However, based on our veterinary experience, older dogs tend to have more abnormalities than younger dogs, highlighting a limitation of the generated documentations. For completeness, the bubble charts for all age bins are presented in the appendix (Figure \ref{fig:bodymap_age_awareness}).

\textbf{Sex.} Since sex was categorized as either male or female, analyzing its impact on stroke distribution was straightforward. However, similar to the age attribute, no differences were observed in the bubble charts or the average number of abnormalities between genders. The results are presented in the appendix (Figure \ref{fig:bodymap_gender_awareness}).

\textbf{Weight.} 
Since weight varies significantly across breeds, we decided to create breed-specific weight bins. For this, we first calculated the weight range (\(R\)) for each breed as:  
\[
R = W_{\text{max}} - W_{\text{min}} \; ,
\]
where \(W_{\text{max}}\) and \(W_{\text{min}}\) represent the maximum and minimum weights for the breed, respectively, as derived from our metadata knowledge base (see Section \ref{sec:metadata}). Next, we divided the range (\(R\)) into three equal intervals to determine the interval size (\(S\)):  
\[
S = \frac{R}{3}
\] 
Using \(S\), we defined three weight bins for each breed as follows:
\begin{align*}
\text{Low}  & : [W_{\text{min}}, \lceil W_{\text{min}} + S \rceil) \\
\text{Medium} & : [\lceil W_{\text{min}} + S \rceil, \lceil W_{\text{min}} + 2S \rceil] \\
\text{High}  & : (\lceil W_{\text{min}} + 2S \rceil, W_{\text{max}}]
\end{align*}
We labeled these bins as \textit{low}, \textit{medium}, and \textit{high}. Again, no clear patterns were observed in the corresponding bubble charts. However, the average number of abnormalities was one higher in the high and low bins compared to the medium bin. Additionally, abnormalities were more frequent in the knee joint and thigh muscles for the heavier bin, though this trend was not consistent across all bins. It is important to note that the minimum and maximum weight data, sourced from Wikipedia, do not account for conditions such as over- or underweight. Considering such factors could strengthen the observed trends and potentially reveal clear patterns. The bubble charts for the three weight bins are shown in the appendix (Figure \ref{fig:bodymap_weight_awareness}).


As shown in Figures \ref{fig:bodymap_age_awareness}, \ref{fig:bodymap_gender_awareness}, and \ref{fig:bodymap_weight_awareness}, the metadata attributes -- age, sex, and weight -- do not appear to significantly influence the generation of synthetic documentations. While the absence of bias concerning a dog's sex is desirable, we expected both age and weight to have a greater impact based on practical veterinary experience. For weight, we identified a potential limitation that may prevent it from having a more significant impact on the generated documentation beyond the minor trends observed.

\subsection{Diversity}
Since each of the 1,000 generated synthetic patellar luxation documentations contains a specific set of abnormalities, each associated with a region and condition index, we wanted to validate whether the LLM produced any duplicates. Detecting duplicate documentations is important, as their presence would not contribute any new knowledge for learning from the synthetic data. Among the 1,000 generated synthetic documentations, we identified six duplicates. These results can be interpreted as an indication that our data generation approach is effective in producing diverse synthetic documentations.

\begin{figure*}[t]
    \begin{minipage}{.48\textwidth}
      \includegraphics[width=\linewidth]{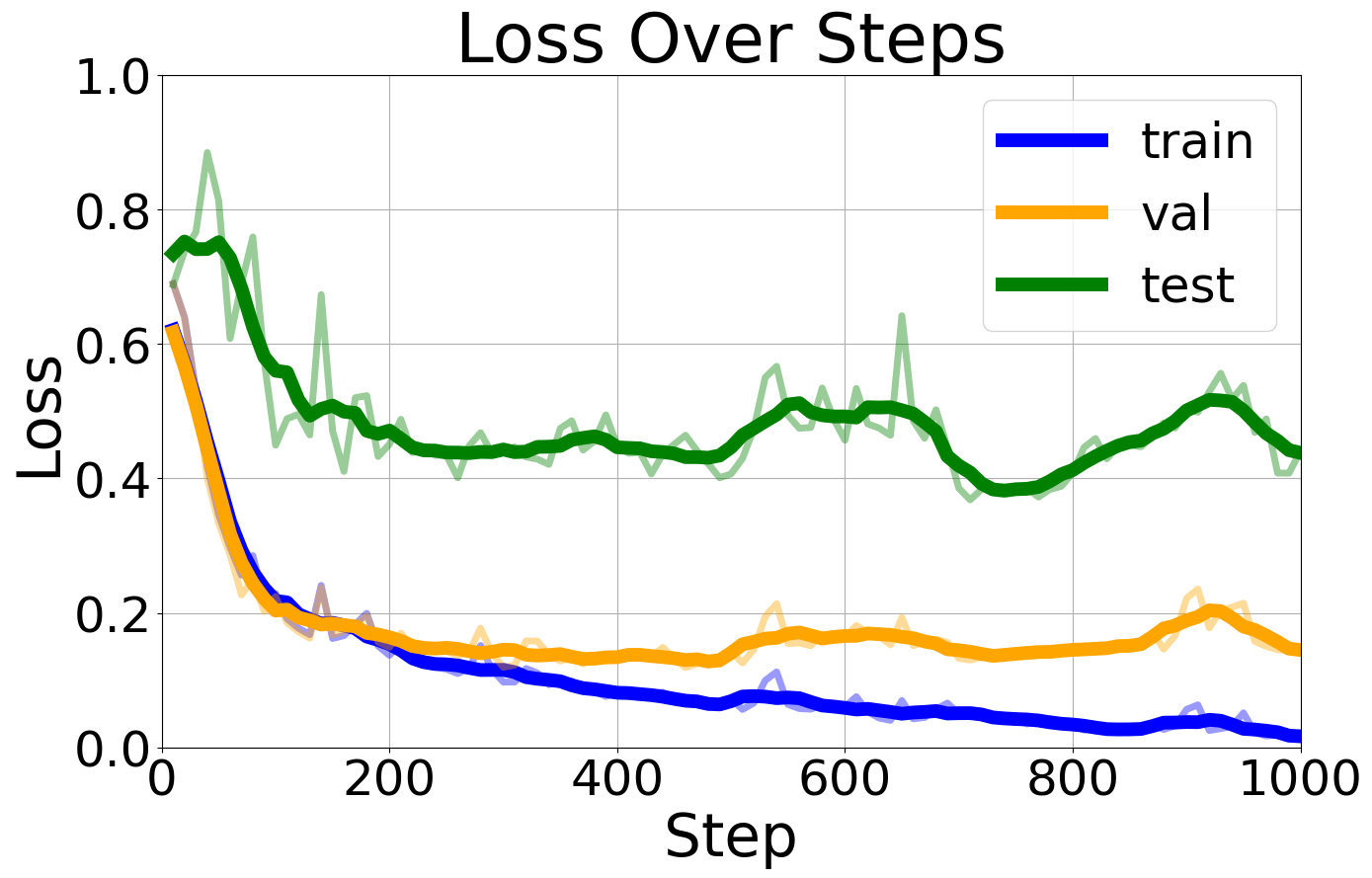}
      \caption{Visualization of the loss curve during fine-tuning the EfficientNet classifier on our generated synthetic data. The curve shows higher loss on the real-world test data, but it generally reflects changes in the validation data as well. After 500 steps, the model appears to begin overfitting.}
      \label{fig:loss_over_steps}
      \vspace{.4cm}
    \end{minipage}
    \hspace*{.02\textwidth}
    \begin{minipage}{.48\textwidth}
      \includegraphics[width=\linewidth]{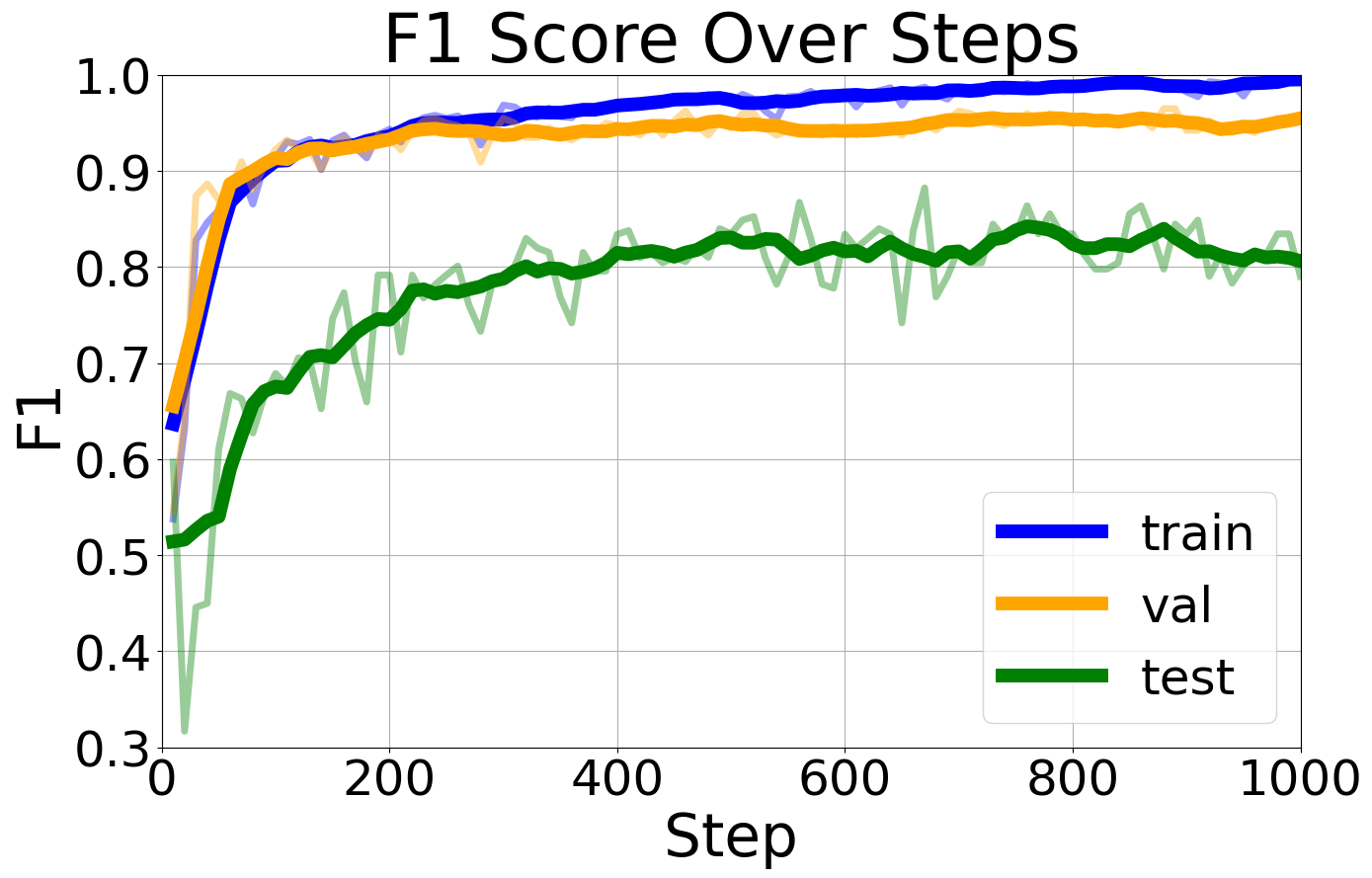}
      \caption{Visualization of the F1 score during fine-tuning the EfficientNet classifier on our generated synthetic data. After 500 steps (5 epochs), the F1 score appears to converge on the test data, consistent with the loss curve observation that the model begins overfitting after 500 steps. The highest F1 score of 88\% on the test data is achieved after 670 steps.}
      \label{fig:f1_score_over_steps}
    \end{minipage}
\end{figure*}

\subsection{Efficiency for Classification Models}
\label{sec:efficiency}
As previously mentioned, we have only 70 real-world documentations available, 19 of which contain a diagnosis of patellar luxation, while the remaining 51 correspond to other diagnoses. To assess the effectiveness of the generated synthetic data for canine musculoskeletal diagnosis, we generated an additional 1,000 synthetic visual documentations across various diagnoses, creating a binary classification dataset to determine whether a documentation represents a patellar luxation or not. This approach allowed us to train a classification model and evaluate its performance on the available real-world data.

Similar to the approach used for generating synthetic documentations of patellar luxation, we explicitly specified the target diagnosis as part of the input prompt to the model. For each run, the diagnosis was sampled from a list of possible conditions derived from the 51 real-world documentations that contain diagnoses other than patellar luxation (e.g., lumbosacral osteoarthritis). Since real-world documentations often include multiple conditions rather than a single diagnosis, we collected a list of 136 distinct diagnoses in total. Using this list ensured that our synthetic data covered the same range of diagnoses as the real-world documentations. However, the LLM was still exposed to only four examples of patellar luxation documentation, as our input prompt remained unchanged, meaning it generated documentations for all 136 diagnoses in a zero-shot setting.

Using 1,000 synthetic documentations per class (patellar luxation and any other diagnosis), we fine-tuned an EfficientNet-B1 model pre-trained on ImageNet \cite{den09}. The 2,000 synthetic documentations were split into 1,600 training and 400 validation samples. Fine-tuning was performed with a batch size of \(16\), a learning rate of \(1 \times 10^{-4}\), and images resized to \(240 \times 240\) pixels for 1,000 steps (10 epochs). The corresponding loss and F1 score curves during fine-tuning are shown in Figure \ref{fig:loss_over_steps} and \ref{fig:f1_score_over_steps}. When evaluated on 70 real-world bodymaps (19 with patellar luxation and 51 with other diagnoses), the model achieved an F1 score of approximately \(88\%\). This indicates that our synthetic data generation approach can effectively support classification tasks, particularly in cases where real-world observations are limited (e.g., rare diseases).

\subsection{Ablation Study}
\label{sec:ablation}
Based on the high F1 score achieved by our classifier, which was fine-tuned solely on the generated synthetic data, we were curious how a classifier trained on a dataset constructed using hand-written rules would perform. In other words, after analyzing 19 real-world patellar luxation documentations, we wanted to test whether the computationally intensive approach of using LLM-generated documentation was necessary, or if a simpler rule-based dataset could achieve similar results. The rules for constructing our baseline dataset are as follows:

\textbf{Patellar Luxation.} The location of the patellar luxation is chosen as either left, right, or bilateral. The condition in the knee joint is then assigned as either chronic or acute. Additionally, with a probability of 20\%, a chronic or acute condition is also assigned to the hip joint on the same side. Between 2 and 8 soft tissue abnormalities are added in the immediate knee area, upper thigh, and lower leg, with the number doubling if the luxation is bilateral. Duplicate abnormalities are not allowed.

\textbf{Other Diagnoses.} For other diagnoses, an integer value is randomly sampled from the range of 3 to 25, representing the number of abnormalities to be added. Each abnormality is then assigned a random region from the 214 possible anatomical regions, along with a corresponding condition selected from seven possible conditions. Duplicate abnormalities are not allowed.

Based on these rules, we created a baseline binary classification dataset comprising 1,000 visual documentations for each class (patellar luxation and any other diagnosis). Additionally, we generated two more binary classification datasets, each modifying our previous approach (see Section~\ref{sec:efficiency}) in a single aspect: In one case, we removed the region information for all 214 anatomical regions from the initial prompt (stated in Section~\ref{sec:generation}); in the other, we removed the four few-shot examples. This allowed us to assess their impact on the generated synthetic data. Using the same methodology as in Section~\ref{sec:efficiency}, our results for all four datasets are presented in Table~\ref{table:ablation_study}.

\begin{table}[ht]
\centering
    \begin{tabular}{lc}
    \hline
    Dataset & F1 Score \\
    \hline
    Full Prompt (LLM) & \textbf{0.88}\\
    \hspace{3mm}w/o Few-Shot Examples & 0.83\\
    \hspace{3mm}w/o Region Information & 0.70\\
    Baseline (Rule-Based) & 0.60\\
    \hline
    \end{tabular}
    \vspace*{0.2cm}
\caption{Results of our ablation study comparing the impact of different synthetic datasets on classification performance.}
\label{table:ablation_study}
\end{table}

Using the full prompt to generate synthetic data produced the best results, achieving an F1 score of 88\%. Removing the four real-world examples from the input prompt led to a slight drop in performance, with an F1 score of 83\%, while removing the region information caused a significant decline to 70\%. The worst performance came from using our simple rule-based baseline dataset, which resulted in an F1 score of 60\%. These findings highlight several key insights. A simple rule-based dataset is insufficient for achieving strong classification performance on the 70 real-world documentation samples. In contrast, synthetic data generated using our proposed approach leads to strong classification results, underscoring its expressiveness. Additionally, adding both few-shot examples and region information enhances the quality of the generated synthetic data and, consequently, improves classification performance. However, specifying possible regions is significantly more important than providing a few real-world examples, which yield only a slight improvement.

\section{Discussion}
The goal of this study was to determine whether LLMs can effectively generate synthetic data for canine musculoskeletal diagnosis. Based on our findings, we believe this question can be answered affirmatively. A lightweight EfficientNet-B1 classification model, fine-tuned on our generated synthetic dataset, achieved a strong F1 score of 88\% when evaluated on 70 real-world documentation samples. In contrast, a model trained on a simple rule-based dataset only achieved an F1 score of 60\%, highlighting both the faithfulness and diversity of the synthetic documentation generated by the LLM. Additional indicators of the synthetic data’s faithfulness and controllability are presented in Section~\ref{sec_awareness}, where we demonstrated that the generated data is sensitive to the given location and grade information of a patellar luxation. In terms of diversity, our analysis of 1,000 generated patellar luxation documentations revealed only six duplicates, highlighting the diversity of the synthetic dataset. Regarding controllability, our findings indicate that while certain attributes -- such as position and severity of patellar luxation -- meaningfully influence the generated documentation, others, including age, sex, and weight, do not appear to have a significant impact. This suggests that, while the generation of synthetic documentation for canine musculoskeletal diagnoses is partially controllable by individual attributes, further research is needed to enhance attribute-specific control. We hypothesize that fine-tuning or incorporating retrieval-augmented generation (RAG) \cite{lew20} could improve the LLM’s awareness of how symptoms and abnormalities vary based on factors such as age, weight, and breed. Specifically, we recommend leveraging specialist literature or research papers in the canine musculoskeletal domain to refine the model's controllability. In addition to our investigation in Section~\ref{sec:efficiency}, further analysis of the optimal number of generated documentations per diagnosis needed for robust classification results could improve efficiency. Expanding our approach to a multi-class classification setting with more real-world test data could further strengthen the validity of our findings. Finally, in contrast to observations reported by the authors of the DeepSeek R1 model \cite{guo25}, we found that including a few example documentations in the input prompt led to improved results (see Section~\ref{sec:ablation}). This suggests that in cases where a standardized output format is desired, providing exemplars within the prompt may be beneficial.

\section{Conclusion}
In this work, we investigated whether LLMs can effectively generate synthetic data for canine musculoskeletal diagnosis, despite their limited prior exposure to this specific use case. Using only synthetic data generated by our proposed method to fine-tune a lightweight classification model, we achieved a strong F1 score of 88\% on 70 real-world test samples. Our findings indicate that certain controllable attributes, such as age and weight, do not meaningfully influence the generated synthetic data, highlighting some limitations. However, we also demonstrated that attributes such as position and grade information of a diagnosis can effectively guide data generation. Additionally, our approach resulted in diverse documentations with minimal duplication. For future work, we suggest that fine-tuning or retrieval-augmented generation (RAG) with high-quality, domain-specific knowledge could improve the influence of controllable attributes like age and weight. Moreover, generating a larger volume of synthetic data for a broader range of diagnoses and evaluating it within a multi-class classification setting with more real-world test data could further validate the promising results presented in this study. These advancements could contribute to improving AI-assisted diagnostic tools in veterinary medicine.


\newgeometry{left=25mm,right=25mm,top=25mm,bottom=13mm,noheadfoot}
\newpage
\onecolumn
\section*{Appendix}
\begin{figure}[h]
    \centering
    \includegraphics[width=\linewidth]{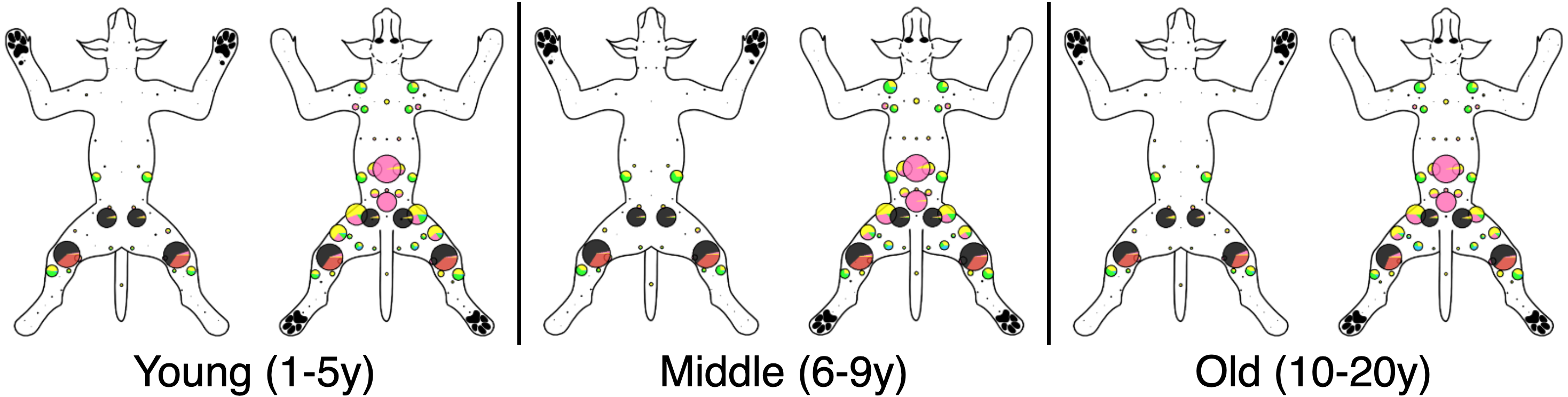}
    \captionsetup{justification=centering} 
    \caption{Bubble charts for all three age bins appear nearly identical, suggesting age does not significantly influence synthetic documentation generation.}
    \label{fig:bodymap_age_awareness}
    \vspace{3mm}
    
    \includegraphics[width=\linewidth]{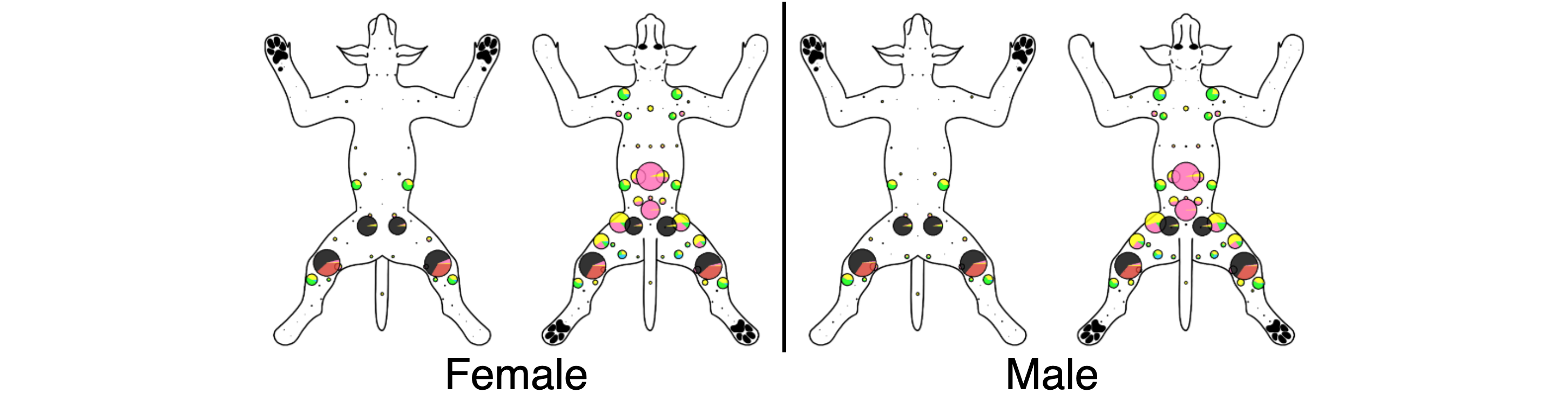}
    \captionsetup{justification=centering} 
    \caption{Bubble charts showing stroke distribution for female and male documentations, with no significant difference observed.}
    \label{fig:bodymap_gender_awareness}
    \vspace{3mm}

    \includegraphics[width=\linewidth]{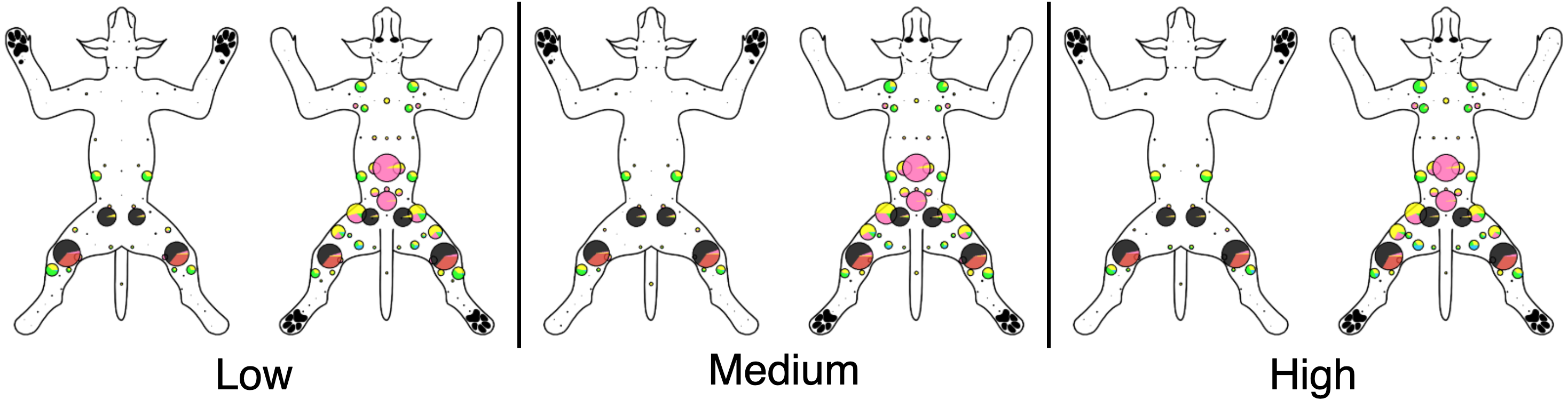}
    \captionsetup{justification=centering} 
    \caption{Bubble charts for the three weight bins show generally larger bubbles in the low and high bins, with the knee joint and surrounding soft tissue having slightly larger bubbles in the high weight bin.}
    \label{fig:bodymap_weight_awareness}
    \vspace{3mm}

    \includegraphics[width=0.6\linewidth]{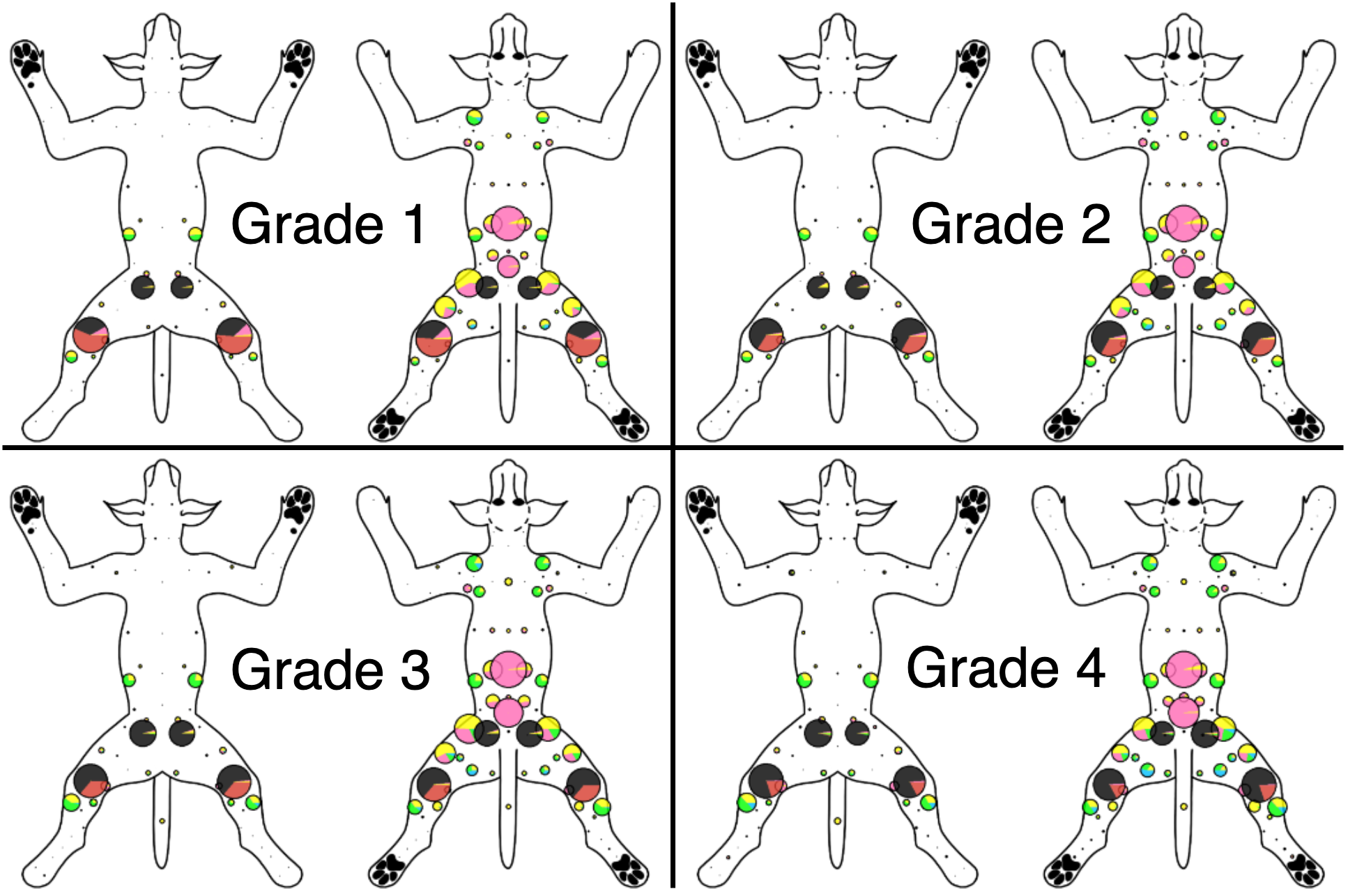}
    \captionsetup{justification=centering} 
    \caption{Higher grades of patellar luxation correspond to more abnormalities. Notably, abnormalities also shift regionally, particularly in the knee joints, where acute inflammation (red) in grade 1 progresses to chronic damage (black) in grade 4.}
    \label{fig:bodymap_grade_awareness}
\end{figure}

\end{document}